\documentclass{article}

\usepackage{arxiv}

\usepackage[utf8]{inputenc} 
\usepackage[T1]{fontenc}    
\usepackage{hyperref}       
\usepackage{url}            
\usepackage{booktabs}       
\usepackage{amsfonts}       
\usepackage{nicefrac}       
\usepackage{microtype}      
\usepackage{lipsum}
\usepackage{graphicx}
\usepackage{amsmath}
\usepackage{multicol}

\usepackage[ruled,vlined]{algorithm2e}

\title{Affect expression behaviour analysis in the wild using spatio-channel attention and complementary context Information}

\author{
  Darshan Gera \\
  \texttt{darshangera@sssihl.edu.in} \\
   \And
S Balasubramanian \\
  \texttt{sbalasubramanian@sssihl.edu.in} \\
}

\begin{document}
\maketitle
\begin{abstract}
Facial expression recognition(FER) in the wild is crucial for building reliable human-computer interactive systems. However, current FER systems fail to perform well under various natural and un-controlled conditions. This report presents attention based framework used in our submission to expression recognition track of the Affective Behaviour Analysis in-the-wild (ABAW) 2020 competition. Spatial-channel attention net(SCAN) is used to extract local and global attentive features without seeking any information from landmark detectors. SCAN is complemented by a complementary context information(CCI) branch which uses efficient channel attention(ECA) to enhance the relevance of features. The performance of the model is validated on challenging Aff-Wild2 dataset for categorical expression classification. Our code is made publicly available\footnote{https://github.com/1980x/ABAW2020DMACS}.
\end{abstract}

\keywords{Facial Expression Recognition \and Spatio-Channel Attention \and Aff-Wild2 \and Efficient channel attention}


\section{Introduction}
FER is an active research area in human-computer interactive systems as expressions convey important cue about emotional affect state of individuals.  Traditional FER systems were built using facial images collected in-a-lab like environment. These methods fail to perform  well under natural and un-controlled conditions of pose, occlusion, lighting etc. To tackle these challenges, large datasets like AffectNet\cite{12} captured in-the-wild have been developed. 

Kollias et al.\cite{kollias2020analysing} as a part of Affective Behavior Analysis in-the-wild (ABAW) 2020 competition have provided benchmark dataset Aff-Wild2 \cite{kollias2018aff, kollias2019expression, kollias2018multi}  consisting of in-the-wild 542 videos with 2,786,201 frames collected from YouTube. This dataset is an extension of Aff-Wild\cite{zafeiriou2017aff, kollias2017recognition, kollias2019deep} dataset. Aff-Wild2 dataset is annotated for 3-different tasks: i) valence and arousal estimation (2-D continuous dimensional model\cite{russell1980circumplex}, ii) seven basic emotions \cite{ekman1992argument} of happiness, neutral, anger, sad, surprise, disgust and fear (categorical classification) and iii) facial action unit recognition based on Facial Action Unit Coding System \cite{ekman2002facial}(multi-label classification). 
Based on our recently proposed end-to-end attention based FER framework\cite{gera2020landmark}, we experiment and validate our model on Aff-Wild2 dataset for expression classification track in ABAW. Our method uses Spatio-channel attention net to capture local and global attentive features along with complementary context information branch using efficient channel attention \cite{26} for obtaining discriminating and robust features in presence of occlusion and pose variations. 

\section{Method}
In this section, we briefly present our proposed framework in \cite{gera2020landmark} for FER under wild conditions of occlusion and pose. The complete architecture is shown in the Figure \ref{fig:figure_attention_fraework}. It consists of 2 branches: i) Local and global attention branch using Spatio-Channel Attention Net and ii) Complementary contextual information(CCI) branch. Next, we discuss each of them briefly:

\begin{figure*}[ht!] 
    \centering
    \includegraphics[width=1.0\textwidth]{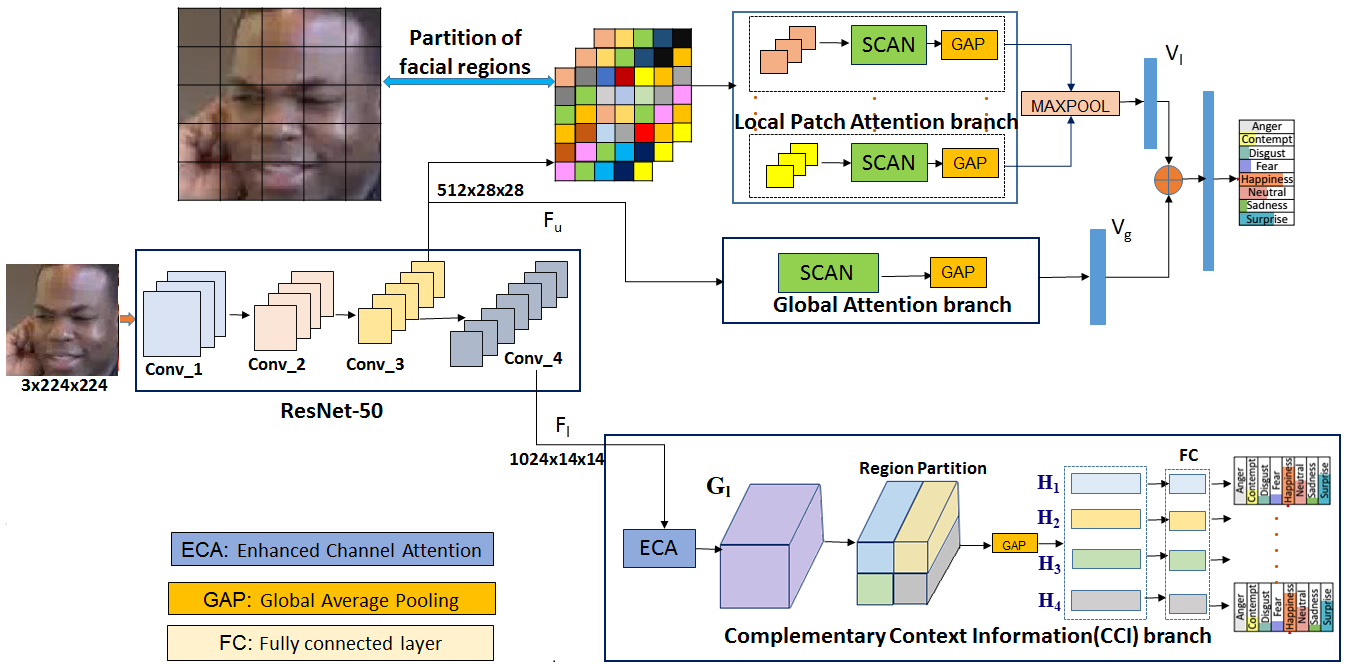}
    \caption{Complete Architecture of our proposed Attention based Facial Expression Recognition Framework\cite{gera2020landmark}.  (Best viewed in color)}
    \label{fig:figure_attention_fraework}
\end{figure*}

\subsection{Spatio-Channel Attention Net(SCAN)}
\begin{figure*}[ht!]
    \centering
    \includegraphics[width=.5\textwidth]{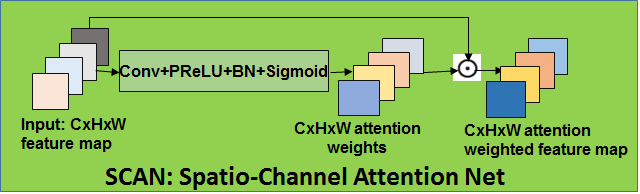}
    \caption{Input output pipeline of SCAN}
    \label{fig:figure_attention_net}
\end{figure*}

 SCAN, an attention block to capture attention per channel per spatial location for the given input feature map was introduced in \cite{gera2020landmark}. The input-output pipeline of SCAN is shown in Fig. \ref{fig:figure_attention_net}. Mathematically, let $I$  be the  $C\times H\times W$ dimensional input feature map to SCAN. Let $f$ be the non-linear function that models SCAN and $W$ be the attention weights of the input feature map computed by SCAN. Then $f$ can be represented mathematically by Eq.\ref{eq:1} as
\begin{equation}\label{eq:1}
    W = SCAN(I) = f(I) = \sigma(BN(PReLU(Conv(I))))
\end{equation}      
where Conv is the ‘same’ convolution operation that takes a $C\times H\times W$ input feature map and outputs $C\times H\times W$ attention weights, followed by parametric ReLU (PReLU) activation, batch normalization (BN) and a sigmoid activation ($\sigma$). 

Following the computation of attention weights, the input feature map is weighted. Let $O$ denote the weighted input feature map. Then,
\begin{equation}\label{eq:2}
     O =  I \odot W
\end{equation}
where $\odot$ is the element-wise multiplication operator. For sake of convenience, it is assumed that this element-wise multiplication is merged with SCAN. Hence the output of SCAN is $O$ for the input $I$. Input $I$ to SCAN could be a local patch from a feature map, or the entire feature map providing global context. For this reason, SCAN is called as local-global attention branch. 

\subsection{Complementary Context Information (CCI) Branch}
While SCAN attends to relevant features locally and globally, extra complementary information is required to enhance the discriminating ability of the model. Transfer learning is used to extract discriminating facial features of eyes, nose, mouth etc. from middle layer of pre-trained model. To eliminate redundancy and emphasize the important features from the middle layer, a channel-wise attention mechanism ECA \cite{26} is applied to the feature map from middle layer. We use ECA not SCAN attention mechanism in CCI branch as it helps to obtain complementary context information to the one provided by SCAN. The CCI branch is similar to the one in \cite{23} but with prior feature emphasis by ECA. Let $F_{l}$ be the feature map from a chosen middle layer of the base model trained for face recognition(FR) and  $G_{l}$  denotes attention feature map after applying ECA on $F_{l}$. let H be the output of CCI branch. Then,
\begin{equation}  \label{eq:3}
      \begin{split}
            H &= CCI(F_{l}) = GAP(PARTITION(ECA(F_{l}))) \\
               &= \{ H_{1}, H_{2}, ..., H_{k}  \} 
      \end{split}
\end{equation}
where PARTITION partitions the ECA weighted feature map into k non-overlapping blocks and GAP is global average pooling. $H_{i}$ is the output of GAP on $i^{th}$ block and it is a vector of features. Here, k is chosen to be 4 based on \cite{23}.

\subsection{Training methodology} 
The input to model is a  $224 \times 224$  RGB image. It is processed by a pre-trained ResNet-50 model\cite{He_2016_CVPR}. For the SCAN branch, the output of $Conv\_3x$ block from ResNet-50 goes as input. This is a $512 \times 28 \times 28$ feature map, denoted by $F_{u}$. Next, $F_{u}$ is partitioned into m non-overlapping blocks $P_{i} (1\leq i \leq m) $, each block acting as a local patch. We have fixed m to be 25 based on ablation studies in \cite{gera2020landmark}. This results in 25 local patches. The output of SCAN from each of the 25 local patches is global average-pooled across spatial dimensions and subsequently max-pooled across channel dimensions to provide the summary of the local context in the form of a 512 dimensional vector, denoted by $V_{l}$. Also, the whole $F_{u}$ is fed to SCAN and the output is global average-pooled to capture global context in the form of another 512 dimensional vector, denoted by $V_{g}$. $V_{l}$ and $V_{g}$ are concatenated and sent through an expression classifier based on cross-entropy loss denoted by $L_{u}$. It is to be noted that the entire local and global context processing through SCAN is devoid of any information from external landmark detectors.

For the CCI branch,  the output of $Conv\_4x$ block from ResNet-50, denoted by $F_{l}$, goes as input. It has dimension  $1024 \times 14 \times 14$. As expressed in Eq. \ref{eq:3}, the output of CCI is a set of k vectors $\{ H_{1}, H_{2}, ..., H_{k}\}$. Subsequently, each of  $H_{i} (1\leq i \leq k) $ is densely connected to a 256 node layer and then goes through an expression classifier based on cross-entropy loss, denoted by $L_{i} (1\leq i \leq k)$. The total loss from the CCI branch is 
\begin{equation} \label{eq:4}
    L_{l} = \sum_{i=1}^{k} L_{i}
\end{equation}
\newline
The overall loss from both the branches is given by
\begin{equation}\label{eq:5}
    L = \lambda*L_{u} + (1-\lambda)*L_{l}
\end{equation}
where $\lambda$ belongs to [0, 1]. Here, $\lambda$ is set to 0.2 based on ablation studies in \cite{gera2020landmark}. 
\newline
\newline
The pseudo code for training the model is shown in Algorithm \ref{Algfer}.

\begin{algorithm}[H]
\SetAlgoLined
 \textbf{INPUT:} dataset(D), parameters($\theta$), model(ResNet-50), number of local patches (m=25), number of blocks(k=4), loss weighing factor($\lambda$) in [0,1] \\
   1. (x,y) $\longleftarrow$ Sampler(D) \\
  2. Obtain $F_{u}$ using Conv\_3x layer of model \\
  3. Obtain $F_{l}$ using Conv\_4x layer of model \\
  4. \textbf{Procedure local-global attentive feature learning}  \\
    \hspace*{1em}i.  Partition $F_{u}$ into m local patches $P_{i} (1\leq i \leq m)$ \\
    \hspace*{1em}ii. Compute local attention features for each $P_{i}$ using SCAN (Eq \ref{eq:1} \& \ref{eq:2}) and after Global average pooling(GAP), perform max pooling to obtain $V_{l}$.  \\
    \hspace*{1em}iii.	Compute global attention features $V_{g}$ for the whole $F_{u}$ using SCAN (Eq \ref{eq:1} \& \ref{eq:2})   \\
    \hspace*{1em}iv.	Concatenate $V_{l}$ and $V_{g}$ and use it to obtain cross-entropy loss ($L_{u}$) \\
  5. \textbf{Procedure complementary context information learning}  \\
    \hspace*{1em}i.	Compute ECA attentive features $G_{l}$ using $F_{l}$ \\
    \hspace*{1em}ii.	Partition $G_{l}$ into k non-overlapping blocks and perform GAP to obtain $H_{i} (1\leq i \leq k)$ using Eq \ref{eq:3} \\ 
    \hspace*{1em}iii.	Obtain cross-entropy loss ($L_{i}$) for each block after passing through FC layer $(1\leq i \leq k)$\\
  6. Obtain combined loss $L$ using Eq \ref{eq:4} \& \ref{eq:5} \\
  7. Update $\theta$ using $L$
 \caption{Spatio-channel attention and complementary context information based learning }
 \label{Algfer}
\end{algorithm}

\section{Datasets and Implementation Details}
\subsection{Datasets}
Aff-Wild2 dataset consisting of 539 videos with 2, 595, 572 frames with 431 subjects, 265 of which are male and 166 female, is annotated for 7 basic facial expressions. Eight of videos have displayed two subjects both of which have been labelled as left and right. Aff-Wild2 is split into three subsets: training, validation and test subsets consisting of 253, 71 and 223 videos respectively. Cropped and aligned frames for all them are made available as a part of the challenge. Since the dataset is highly imbalanced in terms of number of images per expression category, so we use other in the wild datasets like  AffectNet\cite{12}, RAFDB \cite{li2019reliable_45,li2017reliable_46} and ExpW\cite{expw} to make them balanced. \textbf{AffectNet} \cite{12} is the largest facial expression dataset with around 0.4 million images manually labeled for the presence of eight (neutral, happy, angry, sad, fear, surprise, disgust, contempt). Images except contempt category are considered for training the model. \textbf{Expression in-the-Wild (ExpW)} dataset \cite{expw} contains 91,795 images annotated with seven emotion categories. \textbf{RAF-DB} contains 29672 facial images tagged with basic or compound expressions by 40 independent taggers. Only 12271 images with basic emotion from training set are used. Under-sampling and oversampling techniques are also used to overcome imbalance problem.

\subsection{Implementation details}
The proposed work is implemented in PyTorch\footnote{https://pytorch.org} using a single GeForce RTX 2080 Ti GPU with 11GB memory. The backbone net of the proposed architecture is a pre-trained ResNet-50, trained on VGGFace2 \cite{20}.  All images in AffectNet, RAFDB and ExpW are aligned using MTCNN\footnote{https://github.com/ZhaoJ9014/face.evoLVe.PyTorch} \cite{zhang2016_56} and then resized to 224x224. For Aff-Wild2 , cropped images provided by organizers are used after resizing them to 224x224. Input to SCAN is output of Conv\_3x block of backbone Resnet-50.  Input to CCI is output of Conv\_4x block of backbone Resnet-50.  Model is optimized by stochastic gradient descent (SGD) for 20 epochs. Momentum for SGD is 0.9. Learning rate (lr) of backbone is initialized to 0.0001. With regard to SCAN and CCI, lr is initialized to 0.001. lr is reduced by factor of 0.95 every epoch. Batch size is fixed at 64. Weight decay with value 1e-3 is used. Data augmentation used during training consists of random horizontal flipping and color jitter with brightness of 0.4, contrast 0.3, saturation  0.25 and hue of 0.05. To circumvent imbalance dataset, imbalanced dataset sampler\footnote{https://github.com/ufoym/imbalanced-dataset-sampler} is used. The pretrained ResNet50 on VGGFace2 is downloaded from website\footnote{https://github.com/ox-vgg/vgg\_face2}.

\subsection{Evaluation metric}
Evaluation metric used in the challenge evaluation is weighed average of accuracy(33\%) and $F_{1}$ score(67\%). 
Accuracy is defined as fractions of predictions that are correctly identified. It can be written as: 
\begin{equation}\label{eq:6}
Accuracy(Acc) = \frac{Number \hspace{.3em} of\hspace{.3em} Correct\hspace{.3em}Predictions}{Total \hspace{.3em}number \hspace{.3em}of\hspace{.3em} Predictions}
\end{equation}

$F_{1}$ score is defined as weighted average of precision (i.e. Number of positive class images correctly identified out of positive predicted) and recall (i.e. Number of positive class images correctly identified out of true positive class). It can be written as:
\begin{equation}\label{eq:7}
F_{1} \hspace{.3em} score = \frac{ 2 \times  precision \times recall }{ precision + recall}
\end{equation}
And the overall score considered is:
\begin{equation}\label{eq:8}
Overall \hspace{.3em} score = 0.67 \times F_{1} + 0.33 \times Acc
\end{equation}

\section{Results and Discussions}\label{Resultsanddiscussion}
\subsection{Performance Comparison with state-of-the-art methods}
We report our results on the official validation set of expression track the ABAW 2020 Challenge \cite{kollias2020analysing} in Table 1. Our best performance achieves overall score of 0.465 on validation set which is a significant improvement over baseline as well as on many other participating teams performance. For the competition, we have submitted multiple models based i) trained on only train set of Aff-Wild2, ii) trained on both train and validation set of Aff-Wild2 and iii) trained on Aff-Wild2+ExpW+AffectNet .
\begin{table}
\begin{center}
    \caption{Performance comparison on Aff-Wild2 validation set}
    \begin{tabular}{c|c|c|c}
         \hline
         Method & F1 score & Accuracy &  Overall \\
         \hline
         \hline
         Baseline\cite{kollias2020analysing} & 0.21 & 0.664 & 0.36 \\
         ICT-VIPL-Expression\cite{ICT} & 0.333 & 0.64 & 0.434 \\
         NISL2020\cite{NISL} & - & - & - \\
         TNT\cite{TNT} & - & - & - \\
         Ours\cite{gera2020landmark} & \textbf{0.37} & \textbf{0.649} & \textbf{0.465} \\
         \hline
    \end{tabular}
    \end{center}
    \label{tab:Tab1}
\end{table}

\begin{table}
\begin{center}
    \caption{Performance comparison on Aff-Wild2 Test set}
    \begin{tabular}{c|c|c|c}
         \hline
          Method & F1 score & Accuracy &  Overall \\
         \hline
         \hline
         Baseline\cite{kollias2020analysing} & 0.15 & 0.605 & 0.30 \\
         ICT-VIPL-Expression\cite{ICT} & 0.286 & 0.655 & 0.408 \\
         NISL2020\cite{NISL}& 0.27 & 0.68 & 0.405 \\
         TNT\cite{TNT} & \textbf{0.398} & \textbf{0.734} & \textbf{0.509} \\
         Ours\cite{gera2020landmark} & 0.312 & 0.702 & 0.441 \\
         \hline
    \end{tabular}
    \end{center}
    \label{tab:Tab2}
\end{table}

Table 2 presents the performance comparison on test set w.r.t top 3 participating teams. Our top performing model was trained on whole train set of Aff-Wild2 using proposed model which is pretrained on AffectNet, ExpW and RAFDB datasets. Clearly, out model  gives superior performance compared to all teams except TNT\cite{TNT}. They have used both audio and video streams to recognize expressions. Also, their algorithm is heavily dependent on accuracy of landmarks points for detecting the mask whereas our model does not rely on landmarks. 
\begin{table}
\begin{center}
    \caption{Performance comparison w.r.t ResNet50 baseline on validation set}
    \begin{tabular}{c|c|c|c|c}
         \hline
         Method & Pretrained & F1 score & Accuracy &  Overall \\
         \hline
         \hline
         ResNet50 & VGGFace2 & 0.335 & 0.597 & 0.422 \\
        
         Ours & VGGFace2 & \textbf{0.37} & \textbf{0.649} & \textbf{0.465} \\
         \hline
 \end{tabular}
    \end{center}
    \label{tab:Tab5}
\end{table}

\begin{table}
\begin{center}
    \caption{Performance comparison w.r.t number of training samples from Aff-Wild2(using pretrained VGGFace2 model) on validation set. Oversampling is used except for the case when Affectnet and ExpW are used.}
    \begin{tabular}{c|c|c|c|c}
         \hline
         Dataset & Max. no of samples/class & F1 score & Accuracy &  Overall \\
         \hline
         \hline
         Aff-Wild2 & 15K & 0.34 & 0.564 & 0.42 \\
         Aff-Wild2 & 20K & 0.346 & 0.648 & 0.43 \\
         Aff-Wild2 & 50K & 0.35 & 0.635 & 0.447 \\
         Aff-Wild2 & 1 Lakh & 0.348 & 0.642 & 0.445 \\
         Aff-Wild2 & All & \textbf{0.374} & 0.649 & \textbf{0.465} \\
         Aff-Wild2+AffectNet+ExpW & 20K & \textbf{0.379} & 0.616 & 0.457 \\
         Aff-Wild2+AffectNet+ExpW & 50K & \textbf{0.379} & 0.636 & 0.464 \\
         Aff-Wild2+AffectNet+ExpW & 1 Lakh & 0.363 & \textbf{0.664} & 0.462 \\
         \hline
 \end{tabular}
    \end{center}
    \label{tab:tab_number_of_samples}
\end{table}

\begin{table}
\begin{center}
    \caption{Performance comparison on validation set w.r.t different pretrained models. All images of Aff-Wild2 used for training with oversampling.}
    \begin{tabular}{c|c|c|c}
         \hline
         Pretrained model dataset  & F1 score & Accuracy &  Overall \\
         \hline
         \hline
         No-pretrained & 0.249 & 0.493 & 0.329 \\
         AffectNet+RAFDB & 0.353 & 0.615 & 0.439 \\
         AffectNet+ExpW & 0.352 & 0.668 & 0.456 \\
         AffectNet+ExpW+RAFDB & 0.359 & 0.649 & 0.455 \\
         VGGFace2 & \textbf{0.37} & \textbf{0.649} & \textbf{0.465} \\
         \hline
 \end{tabular}
    \end{center}
    \label{tab:Tab7}
\end{table}

\subsection{Ablation studies}
\subsubsection{Comparison with baseline}
ResNet50 pretrained on VGGFace2 dataset with all samples from training set gave overall score of 0.42 as shown in Table 3 but it has large number of parameters as well as training takes much longer time . Also, accuracy is much less as compared to training on our model. 

\subsubsection{Effect of number of training samples}
Since dataset is of huge size and highly imbalanced, we performed a comparative study by training our model on maximum of 15k, 20k, 50k, 1 lakh and all images from Aff-Wild2 as well as by using training images from Affectnet and ExpW datasets. Oversampling is used while training only on Aff-Wild2 dataset. The results are shown in Table 4. Clearly, as number of training data increases, performance improves but gain is not as significant as it should be. By training the model on AffectNet along with ExpW, by using only 50k samples from Aff-Wild2 performance matched to that of by using all images.

\subsubsection{Effect of pretrained dataset}
We did another study by pretraining our model on combination of i) AffectNet + RAFDB, ii) AffecNet+ExpW and iii) AffectNet+ExpW+RafDB and then fine tuned the model on Aff-Wild2 dataset. The results are shown in Table 5. Clearly, pretrained face model is necessary as performance drops significantly without it. Best performance is obtained by using VGGface2 pretrained model.

\subsection{Visualizations}
\begin{figure*}[ht!]
    \centering
    \includegraphics[width=.9\textwidth]{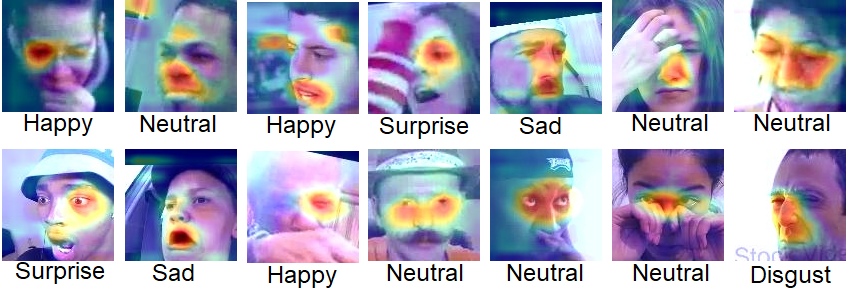}
    \caption{Attention maps for various images from training and validation set of Aff-Wild2 dataset using Grad-CAM activation visualization. (Predicted emotions are shown below each image.)}
    \label{fig:figure_visualization}
\end{figure*}

\begin{figure*}[ht!]
    \centering
    \includegraphics[width=.7\textwidth]{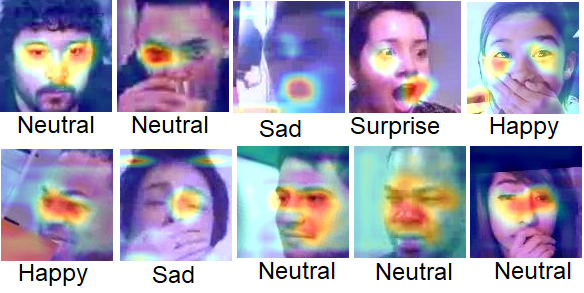}
    \caption{Attention maps for various images from test set of Aff-Wild2 dataset using Grad-CAM activation visualization (Predicted emotions are shown below each image.)}
    \label{fig:figure_visualizationtest}
\end{figure*}

Our attention model is able to capture salient regions in the presence of occlusions and pose variations as shown in Figures \ref{fig:figure_visualization} and \ref{fig:figure_visualizationtest}. These visualizations are obtained using Grad-CAM\cite{selvaraju2017grad}  from Conv\_4x layer feature maps. Dark color indicates high attention whereas lighter color indicates less attention. Figure \ref{fig:figure_visualization} shows visualizations on selected images from train and validation sets of Aff-Wild2 datasets in the presence of occlusion and pose variations. Some of visualizations on images from test set of Aff-Wild2 as shown in Figure \ref{fig:figure_visualizationtest} clearly demonstrate that our model is able to capture non-occluding regions as well as able to handle extreme pose and lighting variations of wild.

\section{Conclusions}
In this paper, we present our spatio-channel attention net and complementary context information  based framework used for facial expression recognition on Aff-Wild2 dataset. Our results on challenging dataset as a part of ABAW 2020 competition demonstrate the superiority of our method as compared to many others methods presented without using any audio and landmarks information. In the future work, we would like to test our model for valence-arousal estimation and facial action unit recognition tasks.

\section{Acknowledgments}
We dedicate this work to Our Guru Bhagawan Sri Sathya Sai Baba, Divine Founder Chancellor of Sri Sathya Sai Institute of Higher Learning, PrasanthiNilyam, A.P., India. We are also grateful to D. Kollias for all patience and support.

\bibliographystyle{unsrt}  
\bibliography{references}

\end{document}